\documentclass[11pt,a4paper]{article}
\usepackage[hyperref]{acl2020}
\usepackage{times}
\usepackage{latexsym}
\usepackage{balance}
\usepackage{stfloats}
\usepackage{tabularx}

\usepackage{soul}
\usepackage{microtype}

\aclfinalcopy 
\def\aclpaperid{14} 

\title{Leveraging Multimodal Behavioral Analytics for Automated Job Interview Performance Assessment and Feedback}

\author{
Anumeha Agrawal \thanks{equal contribution} ,
Rosa Anil George \footnotemark[1] , 
Selvan Sunitha Ravi \footnotemark[1] ,
Sowmya Kamath S and \\
\textbf{Anand Kumar M}
\\
Department of Information Technology\\
National Institute of Technology Karnataka, Surathkal, India 575025\\
\{anumehaagrawal29, rosageorge97, sunitha98selvan\}@gmail.com,\\
\{sowmyakamath, m\_anandkumar\}@nitk.edu.in 
}

\date{}

\begin{document}
\maketitle
\begin{abstract}
Behavioral cues play a significant part in human communication and cognitive perception. In most professional domains, employee recruitment policies are framed such that both professional skills and personality traits are adequately assessed. Hiring interviews are structured to evaluate expansively a potential employee's suitability for the position - their professional qualifications, interpersonal skills, ability to perform in critical and stressful situations, in the presence of time and resource constraints, etc. Therefore, candidates need to be aware of their positive and negative attributes and be mindful of behavioral cues that might have adverse effects on their success. We propose a multimodal analytical framework that analyzes the candidate in an interview scenario and provides feedback for predefined labels such as engagement, speaking rate, eye contact, etc. We perform a comprehensive analysis that includes the interviewee's facial expressions, speech, and prosodic information, using the video, audio, and text transcripts obtained from the recorded interview. We use these multimodal data sources to construct a composite representation, which is used for training machine learning classifiers to predict the class labels. Such analysis is then used to provide constructive feedback to the interviewee for their behavioral cues and body language. Experimental validation showed that the proposed methodology achieved promising results.
\end{abstract}


\section{Introduction}

In the business world, interviews are a prerequisite to personnel recruitment for assessing the candidates through a structured interaction and discussion either on a one-to-one basis or by a panel of interviewers. It is an opportunity for the candidates to prove that they are qualified for the position, and for recruiters to assess the job-to-candidate fit. Such recruiters are trained in evaluating a candidate's personality, thought patterns, behavior under stressful situations, and emotional intelligence through well-established metrics through technical analysis, psychometric testing, etc. Several theoretical models  have suggested the ``big five trait taxonomy", based on which an individual's traits can be summarized, and scoring can be performed for choosing a candidate \cite{john1999big}. 

The field of personality computing focuses on automatically analyzing such essential insights into the psyche of a person based on their behavior, verbal responses and non-verbal actions, speech patterns, body language, etc \cite{vinciarelli2014survey}. Studies have shown that nonverbal behaviors such as smiling, maintaining eye contact, and good posture all contribute significantly to interpersonal communication along with an indication as to the mental health and well-being of the participants \cite{2019novel}. The significant difference between verbal and non-verbal communication is that the former is specific and interpreted, while the latter is subtle and implied. Both channels of communication affect conversational dynamics and influence the relationship between individuals \cite{Vikrant:2015:RES:2877810.2877854}.  

Automating such behavioral analysis can be beneficial while conducting mass recruitment drives, primarily when many of these are done online through videoconferencing systems. From the candidates' perspective, knowing the effect of their own verbal and non-verbal behavior in creating a favorable impression and increasing their chances of success in interviews is also of significant interest. These cues may be available through their interview videos and audio, speaking qualities, and the effect of content delivery. Analysis of a single modality can often be insufficient in obtaining usable insights into how humans perceive and express information \cite{arsnitk}. Modalities other than text can commonly present clues for the expression of sentiment and feelings \cite{ding2013360}. Audio and visual features also aid in linguistic disambiguation as they provide additional details regarding the speakers' sentiment. When a person speaks with optimal vocal modulation or using appropriate gestures, it conveys a lot more than just the content. 

Research on automated analysis of both verbal and non-verbal behavior cues in the case of job interviews has recently gained momentum, which we aim to analyze in this work. In this paper, we experiment with several linguistic, audio, and visual features extracted from recorded interviews to create feature vectors passed to different classifiers to score the candidate on specific predefined labels. The remainder of this paper is organized as follows: Section 2 presents a discussion on existing works in the area of interest. The details of the proposed methodology and specifics of implementation are presented in Section 3. In Section 4, the experimental analysis and performance of the various models for each task are discussed, followed by the conclusion and references.

\section{Related work}

Over the last decade, behavioral patterns analysis using multimodal data has received significant research attention. \citet{navas2006objective} conducted experiments for speech-based emotion analysis to compare speaker-dependent and speaker-independent techniques. To perform this analysis, several acoustic features such as fundamental frequency, duration, intensity are extracted to find hidden information. However, the speaker-dependent approach was not scalable and cannot be used in large-scale applications with several users. \citet{Borth:2013:LVS:2502081.2502282} proposed a different approach for sentiment analysis of visual content using SentiBank. SentiBank is a visual concept detector library used to extract various concepts and Adjective Noun Pairs (ANP) from the visual content. While existing models predicted sentiments or emotions directly from low-level visual features, their approach used high-level visual features to better capture sentiments. They use images to extract mid-level semantic features and use a classifier to predict semantic features, which can be used to determine the relevance and importance of the image in determining emotion. 

\citet{nguyen2013multimodal} used real-time interview data to monitor and analyze body communication cues. The interviewees are seated in the videos, which lets them analyze both upper body movement and facial cues. Various visual features are automatically extracted, and data is annotated to predict the personality and job interview ratings. This model shows the importance of using bodily gestures to predict the personality and give interview ratings. \citet{naim2015automated} collated and used the MIT Interview dataset and trained Lasso and SVR models to predict several emotions and verbal/nonverbal cues like EyeContact, Calm, Speaking Rate, Authentic, Focused, Structured Answers, Smiled, Friendly, Engagement, etc. These labeled ratings were manually assigned by Amazon Turk Workers, and the ground truth labels were derived by averaging the scores of 9 Turk workers. 

\citet{pereira2016fusing} presented a new technique for sentiment analysis in the telecommunication domain. They extracted and combined prosodic, lexical, and visual features from news videos and applied various computational methods to recognize real-time emotions from facial cues. The speech delivered by each participant is processed, parsed, and sentiment analysis is done on the corresponding text transcript. Features such as visual power of perceived emotion, field sizes of members, voicing likelihood, sound intensity, the fundamental frequency of the speech, and the scores associated with the sentiment were defined and used. One of the limitations is that the poor audio quality of the chosen dataset resulted in inaccurate sentiment prediction. Another drawback observed is due to the selected distance metric, the model does not map well to the intensity of the sentiment.

In recent years, behavioral pattern analysis using multimodal data has received significant research attention. The rise in online video streaming and hosting websites such as YouTube has facilitated an increase in sentiment expressions in multiple modalities \cite{sentiment2017domain}. The availability of standard datasets containing videos annotated for emotion or sentiment has also has been conducive to this, as shown in \cite{zadeh2016multimodal}. 

\citet{chen2016automatic} generated a multimodal corpus with structured interview responses, by manually rating the interviewee’s personality and performance for 12 structured interview questions which measure different types of job related skills. Along with interviews, the interviewee's public speaking videos were also recorded and used to provide useful cues. They used visual, lexical and speech features, based on which they showed that using both non-verbal and verbal cues outperforms other cases. \citet{poria2017context} studied the emotions from facial expressions, reporting that standard facial expressions are sufficient to provide several clues to detect emotions. Speech-based emotion analysis based on the identification of various acoustic features, such as the intensity of utterance, bandwidth, pitch, and duration, is also helpful.  
They achieved a 5-10\% improvement in performance compared, however, the contextual relationship between utterances are considered and treated equally in this model. \citet{cambria2017benchmarking}'s multimodal emotion recognition model extracts features from text and videos using a convolutional neural network architecture, incorporating all three modalities- visual, audio, and text. \citet{radhakrishnan2018sentiment} proposed a new approach for sentiment analysis from audio clips, which uses a hybrid of the Keyword Spotting System. The Maximum Entropy classifier was designed to integrate audio and text processing into a single system, and this model outperformed other conventional classifiers.
 
 \citet{blanchard2018getting} developed a fusion technique for audio and video modalities using audio and video features to analyze spoken sentences for the sentiment. They did not consider the traditional transcription features to minimize human intervention. However, the model can be scaled and deployed in the real world effortlessly. They selected high dimensional features for the model to test their generalizability in the sentiment detection domain. \citet{Hu:2018:MSA:3219819.3219853} presented a novel approach that uses deep learning to identify the sentiment of multimodal data. The modalities considered were images and text, and computer vision techniques were combined with text mining. Their aim was to treat it as a study of emotion, one of the most exciting fields in psychology. They did this using a large social media dataset of Tumblr posts, using which the emotion word tags attached by users was predicted, treating these as emotions reported by the user. Their work combined image and text and proved that combining these two modalities conveyed more information about the sentiment that either of the modalities alone.  

Based on the review of existing work, several limitations were observed. When features from different modalities are considered, it is crucial to find only those features that influence the label. Thus, we aim to address the issue of feature selection by experimenting with different feature selection algorithms. Many features are strongly correlated to each other, and considering these strongly correlated features together will not add a lot of value in predicting a label. Identifying and removing such features that are strongly correlated to each other and considering only one such feature in predicting the label can be more beneficial.  

\section{Proposed Approach}
In this section, the proposed model and its associated processes are described in detail. Our approach is built on all three modalities, and the data modality-specific preprocessing techniques and various algorithms used to classify the data for each of the labels, are presented. For experimental validation, we used the MIT interview dataset \cite{naim2015automated}, which consists of recordings of 138 mock job interviews of 69 candidates pre-intervention and post-intervention. It contains Amazon Mechanical Turk Worker scores for each video, which when averaged gives the final score for each of the labels. It includes the audio files as well, which we use for audio analysis.  In contrast to \citet{naim2015automated} who used regression as their evaluation metric, we use classification, with the numbers 1 to 7 representing the level of performance. A score of 1 for any label is treated as very bad performance whereas a score of 7 is treated as exceptionally good. We use class labels as, our objective is to provide users with feedback based on the class label.

\subsection{Prosodic Features}
Prosodic features play an essential role in characterizing the speaking style of the interviewee. Frequency, pitch information, tone, intensity, spectral energy, spectral centroid, zero-crossing rate, etc. are some of the prosodic features which are considered to be primary in analyzing the speaking style and emotions. In \cite{naim2015automated}, the pitch information, vocal intensities, characteristics of the first three
formants, and spectral energy were included. We found that time-domain features are of utmost importance as the emotion can be predicted by considering several frames together. We used the raw audio signals to extract the time domain features. Using the magnitude of the Discrete Fourier Transform (DFT), we calculated the frequency domain features. The cepstral domain is computed using the Inverse DFT on the logarithmic spectrum. These features can be extracted for windows of small and large sizes. In our methodology, we used a short term window and split the audio signal into short-term windows (frames). We extracted features for each frame, giving us a feature vector of 40 elements, using a short term window size between 20ms and 100ms. The pyAudioAnalysis \cite{giannakopoulos2015pyaudioanalysis} library of Python was used to generate the features for the audio. Other frameworks, such as PRAAT \cite{boersma2018praat}, can also be used to extract prosodic features. 

\subsection{Facial Expressions}
Landmarks are to be first captured for extracting facial features, which are essential points of interest on a person's face. The global transformations, including rotation, translation, and scaling were disregarded, and only the local changes were considered while extracting features from the tracked interest points. These local changes can provide useful information about our facial expressions. OpenCV was used to extract each landmark, namely, nose, left mouth, right mouth, chin, left eye left corner, and right eye right corner from each frame. The video was broken down into frames of size 1 second, and features are extracted. These were averaged over the given time frame of the video. We also incorporated the head pose features (Pitch, Roll, and Yaw) based on the corresponding elements of the rotation matrix $R$. A pre-trained convolutional neural network called LeNet \cite{lecun1998gradient}, consisting of two alternate Conv layers, a pooling layer, and finally, a fully connected layer was used for detecting a person smiling or not was used. LeNet was trained on the SMILES dataset \cite{arigbabu2016smile} consisting of 13,165 face images, of dimension 64x64x1 (grayscale). 

\subsection{Lexical Features}
The linguistic features provide insightful information regarding the confidence and the style of speech of the interview candidate. The most commonly used feature for text is the counts of individual unique words. It gives a clear understanding of proficiency, eloquence, and the ability to use proper vocabulary during a structured communication episode like an interview.

To obtain lexical features, the text transcripts of all the audio clips were obtained, using the Google Cloud Speech-to-Text API. Once we obtain all the text transcripts, text cleaning was done before further processing. {\color{black} All letters were converted to lowercase, after which punctuation marks, accent marks, and any extra white spaces were removed. Tokenization was performed to split the text into smaller units, using the Natural Language Toolkit \cite{bird2008nltk}, a Python library for tokenization}. Next, the speaking style features were extracted, like the average number of words spoken per minute, the average number of unique words per minute, count of unique words in the transcript, and the number of filler words used per minute. Information regarding speaking rate, proficiency, and fluency of a particular candidate can be evaluated using these features. \citet{pereira2016fusing} computed the sentiment score for each sentence in the closed captions as a summation over the generated vector assigning the sentiment (-1,0,1) for each method. We incorporated a similar logic to obtain the emotion scores. 

Finally, to get a detailed analysis of the overall emotion of the text, we used the Tone Analyzer \cite{akkiraju2015ibm}. Each sentence is passed through the Tone Analyzer, and the percentage of emotion in that sentence in the following categories- Joy, Sadness, Tentative, Analytical, Fear, and Anger, is calculated. Each interview is assigned the average score per category, as mentioned earlier. The Stanford Named Entity Recognizer (NER) \cite{finkel2005incorporating} was also used to obtain the count of nouns, adjectives, and verbs in each sentence.

\subsection{Class Prediction}
Several experiments were carried out after and before the feature selection process. Each of these experiments was carried out for individual labels, and we made several interesting observations. We experimented with four machine learning algorithms - Random Forest, Support Vector Machine Classifier (SVC), Multitask Lasso Model, and Multilayer Perceptron (MLP). 

Random Forest \cite{10.1023/A:1010933404324} build decision trees on data samples that get chosen randomly. A prediction is obtained from each decision tree, and the most optimal solution is selected through voting.  We used attribute selection techniques such as Information gain, Gini index, and Gain ratio to generate each decision tree and obtain the final voting. With this model, the issue of overfitting is avoided as the biases get canceled out by taking an average of the predictions. Support Vector Classifiers \cite{10.1023/A:1022627411411} find an appropriate hyperplane in an N-dimensional space to classify the data points distinctly. The support vector classifier aims to maximize the margin between the hyperplane and data points. The Multi-task Lasso model \cite{10.5555/3042573.3042652} penalizes least-squares along with regularization to suppress or shrink features. The Lasso makes use of both feature selection and continuous shrinkage due to the nature of the norm penalty. The optimization objective for Lasso can be calculated using Eq. (\ref{eq1}) and Eq. (\ref{eq2}), where $n$ represents the sample size considered, $Y$ is the vector containing the target values, $X$ is the training data, $W$ denotes the weight matrix, $\alpha$ is a constant that is multiplied with the L1-norm of the coefficient vector.  
    \begin{equation}
        1 / (2 * n_{samples}))*||Y - XW||^2     
        + \alpha ||W||_{21} \label{eq1}
    \end{equation}
    where, 
    \begin{equation}
        ||W||_{21} = \sum_i \sqrt{\sum_j w_{ij}^2}  \label{eq2}
    \end{equation}

The Multi-layer Perceptron (MLP) is a supervised learning algorithm that helps the target learn a non-linear function approximator, given a set of features. There may exist one or may non-linear layers known as hidden layers between the first (input) and last (output) layers. A single hidden layer makes the model a universal approximator, while also supporting multi-label classification and learning non-linear models. 

\subsection{Implementation}
The features generated from the three different modalities - text, audio, and video were used to construct a feature vector, which is then passed through various classifiers to predict the class for the labels. As discussed, four different algorithms - Random Forest, SVC, Multitask Lasso, and MLP were used as classifiers. The features used to build the feature vector were: 
\begin{enumerate}
    \item \textit{Audio} - Power, intensity, duration, pitch, zero-crossing rate, energy, the entropy of energy, spectral centroid, spectral flux, spectral spread, spectral roll-off, MFCCs, Chroma vector, Chroma deviation.
    \item \textit{Video} - Nose, chin, left eye left corner, right eye right corner, left mouth, right mouth landmarks, yaw, pitch, roll, smiling or not smiling. 
    \item \textit{Lexical} - speaking rate, proficiency, fluency, count of total words spoken, Count of total unique words spoken, the emotion of the text, the score associated with the emotion, count of Nouns, Verbs, Adjectives.
\end{enumerate}

Once the fused feature vector is obtained, the ML classifiers are trained to predict the ratings of the interview on a scale of 1-7 based on 9 different parameters - Eye Contact, Speaking rate, Engaged, Pauses, Calmness, Not stressed, Focused, Authentic and Not Awkward. These parameters are influenced by a set of selected features from the feature vector, as processed from the dataset. Thus, we take different combinations of lexical, prosody, and facial features to find the optimal features for each of the parameters. Feature selection techniques are also employed for obtaining the optimal feature vector. The data is first normalized using standardization and scaled to unit variance. The standard score of sample x is calculated using the Eq. (\ref{eq3}), where $u$ is the mean of the training samples, and $s$ is the standard deviation of the training samples. 
\begin{equation}
 z = (x - u) / s\label{eq3}
\end{equation}

Each feature is centered and scaled based on the mean computed for the samples in the training set. Automatic feature selection is carried out to eliminate redundant and irrelevant features. Different feature selection processes are performed for each of the parameters. During K best feature selection, a correlation matrix is calculated, and $k$ features that have the highest scores indicating strong relationships with the output variable are retained, while the other features are eliminated. 

Based on the feature vector selected, we also experimented to find the optimal value of k as well. We used the Benjamini-Hochberg procedure \cite{thissen2002quick} to decrease the false discovery rate, as it helps control the influence of small $p$-values, which often leads to rejection of a true null hypothesis. Due to this, the number of false positives is mostly decreased. For this, the $p$ values for all variables are calculated and then ranked. The variables with $p$ values higher than a threshold value are retained, while all other variables are eliminated. Family-based errors are used to calculate the probability of false positives so that features that cause Type I errors can be eliminated.

During our experiments, we found that several features were correlated with each other. In such cases, just one of the features can be retained, and the rest can be ignored. Through several experiments, we determined the ideal threshold value for correlation, as 0.6. The ML models were then trained to predict ratings for each of the label parameters. We try different combinations of feature selection methods and algorithms and observed their effect on the performance. This helped in understanding the most well-suited model for different settings. For each algorithm, an extensive search is performed over specified hyper-parameter values to help ensure that the models do not perform poorly due to a lack of hyperparameter tuning. We used 3 fold cross-validation to ensure that our models perform well in the real world as well. 

\begin{table*}[ht]
\centering
\setlength{\tabcolsep}{8pt} 
\renewcommand{\arraystretch}{1}
\caption{\label{table-1}Best accuracy scores obtained for models trained on audio+video+lexical multimodal feature vector}
\begin{tabular}{lcccc}
\hline
\textbf{Label}  & \textbf{Random Forest} & \textbf{SVC} & \textbf{Multitask Lasso} & \textbf{MLP}\\
\hline
Eye contact & 0.5714 & 0.5714 & 0.5714 & 0.5714\\
Speaking rate & 0.9643 & 0.8928 & 0.9642 & 0.7857\\ 
Engaged & 0.6428 & 0.7500 & 0.6428 & 0.5383 \\
Pauses & 0.8214 & 0.7857 & 0.6785 & 0.6785 \\
Calmness & 0.7857 & 0.7857 & 0.7500 & 0.6071 \\
Not Stressed & 0.8214 & 0.8214 & 0.6785 & 0.7500 \\
Focused & 0.7500 & 0.7142 & 0.7142 & 0.7142 \\
Authentic & 0.6787 & 0.6428 & 0.4642 & 0.6428\\
Not Awkward & 0.6071 & 0.464 & 0.4285 & 0.45357\\
\hline
\end{tabular}
\end{table*}

\section{Experimental Results and Analysis}
For extensively evaluating the proposed multimodal analytics pipeline, various combinations of prosodic, visual, and lexical features were experimented with, and used to train the four different classifiers, discussed in Section 3. each classifier is trained to predict from 9 different class labels - \textit{eye contact, speaking rate, engaged, pauses, calmness, not stresses, focused, authentic, and not awkward.} 
Experiments using the fused multimodal feature vector were performed, the results of which are tabulated in Table 1. As can be observed from Table 1, the Random Forest Classifier outperformed the others for the \textit{Eye Contact} class, with an accuracy of 64.28\% obtained when the family-wise error technique of feature selection was used (further experiments were conducted to evaluate the effect of different modalities, this is presented in Table 2). Most models were able to predict a rating for \textit{Speaking rate} with high accuracy of 96.43\%. The Lasso Classifier with Benjamini-Hochberg technique and Random Forest Classifier with the family-wise error technique helped achieve the best results. For the \textit{Engaged} label, an accuracy of 75\% using the Support Vector Classifier, along with the family-wise error technique of feature selection was obtained, while for \textit{Pauses}, an accuracy of 82.14\% using the Random Forest Classifier and the K best feature selection was seen as the best.

Two models performed well on the dataset to achieve an accuracy of 78.57\% on the \textit{Calmness} parameter - the Support Vector Classifier with the Benjamini-Hochberg technique and the Random Forest Classifier with K best feature selection technique. The Random Forest Classifier and Support Vector Classifier achieved an accuracy of 82.14\% for \textit{Not Stressed} label, while the Random Forest Classifier with the family-wise error technique outperformed other variations for the \textit{Focused} label. For the \textit{Authentic} label, the best accuracy obtained was only 67.87\% using the Random Forest Classifier with the family-wise error technique of feature selection when Lexical and Facial features were used. For the \textit{Not Awkward} label, the Random Forest Classifier showed the best performance at 60.71\%, though still low when compared to other class labels. Random Forest performed the best for 8 out of 9 labels. This is because it selects features that contribute the most to the classification as it considers the average of all predictions, canceling out the bias. 
However, MLP underperformed on most of the parameters.

\begin{table*}[ht]
\centering
\setlength{\tabcolsep}{7pt} 
\renewcommand{\arraystretch}{1}
\caption{\label{table-2}Observed accuracy scores for different combinations of modalities using a Random Forest Classifier.}
\begin{tabular}{lccccc}
\hline
\textbf{Label} & \textbf{Audio+Video+Lexical} & \textbf{Audio+Video} & \textbf{Lexical+Video} &\textbf{Audio+Lexical}\\
\hline
Eye Contact & 0.5714 & 0.6428 &	0.5714 &	0.5428 \\
Speaking rate & 0.9643 & 0.9285 &	0.9285 &	0.9285 \\
Engaged & 0.6428 & 0.6428 &	0.6071 &	0.6071 \\
Pauses & 0.8214 & 0.7857 &	0.7857 &	0.7857 \\
Calmness & 0.7857 & 0.7500 &	0.7500 &	0.7857 \\
Not Stressed & 0.8214 & 0.7500 &	0.7142 &	0.8214 \\
Focused & 0.7500 & 0.5714 &	0.6071 &	0.6071 \\
Authentic & 0.6787 & 0.6071 &	0.6785 &	0.6428 \\
Not Awkward & 0.6071 & 0.4285 &	0.2500 &	0.4285 \\
\hline
\end{tabular}
\end{table*}

\begin{table*}[ht]
\centering
\setlength{\tabcolsep}{12pt} 
\renewcommand{\arraystretch}{1}
\caption{\label{table-3} Observed accuracy scores for individual modalities using Random Forest Classifier}
\begin{tabular}{lccc}
\hline
\textbf{Label} & \textbf{Audio} & \textbf{Video} & \textbf{Lexical}\\
\hline
Eye Contact &	0.5000	& 0.6071 & 	0.4642 \\
Speaking rate & 	0.8571 & 	0.8571 & 	0.8571 \\
Engaged	& 0.5357 & 	0.4642 & 	0.5357\\
Pauses	& 0.6071 &	0.6071 &	0.6428 \\
Calmness &	0.6785 &	0.6785 &	0.6785 \\
Not stressed & 0.7500 & 0.7142 & 0.7857 \\
Focused & 0.7857 & 0.7142 & 0.6785 \\
Authentic &	0.4642 & 0.5357 & 0.5714 \\
Not Awkward	& 0.5000 &	0.2857 &	0.2142\\
\hline
\end{tabular}
\end{table*}

\begin{table*}[ht]
\centering
\setlength{\tabcolsep}{7pt} 
\renewcommand{\arraystretch}{1}
\caption{\label{table-4}Best accuracy scores for different feature selection techniques.}
\begin{tabular}{lccccc}
\hline
\textbf{Label} & \textbf{Benjamini-Hochberg} & \textbf{Family-wise error selection} & \textbf{K best feature selection}\\
\hline
Eye Contact & 0.5714 & 0.6428 &	0.5714\\
Speaking rate & 0.9643 & 0.9643 &	0.9285  \\
Engaged & 0.6428 & 0.7500 &	0.6071  \\
Pauses & 0.7124 & 0.7857 &	0.8214  \\
Calmness & 0.7857 & 0.7500 &	0.7857  \\
Not Stressed & 0.7142 & 0.8214 &	0.7857  \\
Focused & 0.6875 & 0.7500 &	0.6071  \\
Authentic & 0.5357 & 0.6787 &	0.6428 \\
Not Awkward & 0.6071 & 0.5714 &	0.5357  \\
\hline
\end{tabular}
\end{table*}

Another objective was to check how the different modalities measure up when feature sets using any two modalities are created and ML models are trained using these features. This basically provides insights into which modalities provide an edge in capturing personality-specific traits. To assess this behaviour, we conducted experiments with three combinations of modalities - \textit{Audio+video}, \textit{Lexical+video} and \textit{Audio+lexical} as well as experimented with individual modalities. The random forest classifier was trained on feature sets generated by fusing these modalities to create three different two-modality feature sets, after which the label prediction performance was then observed. Similarly, we also considered each of the three modalities on their own, that is, the audio, video and lexical feature sets. Again, the best performing classifier, Random Forest was trained separately on the one-modality feature vectors, and label prediction performance was observed. 

Table 2 and Table 3 show the results obtained from two-modality feature vectors and each individual modality feature vector for the best performing classifier, Random Forest, respectively. We observed that the performance varies significantly when the classifiers are trained on different combinations of the modality-specific feature set. For the \textit{Eye contact} class, the \textit{Audio+video+lexical} feature vector was not very accurate. In fact, the two-modality feature vector performed better than the three-modality feature vector. However, for all other classes, the Random Forest classifier trained on the three-modality feature vector outperformed all other variants. Table 4 shows the accuracy scores for the different feature selection techniques when all the modalities are considered.

\section{Concluding Remarks}
In this paper, approaches to automatically assess candidates' strengths and weaknesses during an interview, using the video, audio, and transcripts of the interview was presented. Various preprocessing steps, including normalization and feature extraction for each of the three modalities was performed, followed by feature selection to select the best features from each modality. Label classification was performed using four machine learning models - Random Forest, Support Vector Classifier, Multi-task Lasso, and Multi-Layer Perceptron model on the optimal set of fused features and their variations. Effect of various combinations of modalities and feature selection techniques are experimented with. The models were trained for prediction with respect to nine labels to evaluate the candidate. Experiments revealed that the Random Forest Classifier outperformed all other models for 8 out of the 9 labels considered. 

The current dataset has only 169 videos, making it difficult to get a very high accuracy for all the labels. The dataset could be expanded to include more interview videos that are scored by Amazon Turk workers. We also aim to improve the predictions by incorporating behaviors such as hand movements and body posture to get a refined understanding of the candidate's performance. The current model will be integrated into a web application that can be used as a feedback tool to train candidates for interviews by providing them with real-time feedback on their performance and pointers to manage their strengths and weaknesses. The scores can then be interpreted to give meaningful suggestions to the candidate for boosting their interview performance.

\balance
\bibliographystyle{acl_natbib}
\bibliography{acl2020}
\end{document}